# Learning Machines: In Search of a Concept Oriented Language

Veyis GUNES

*Independent researcher, Le Mans, France*
*e-mail: v.gunes@laposte.net*

*August 20, 2018*

What is the next step after the data/digital revolution? What do we need the most to reach this aim? How machines can memorize, learn or discover? What should they be able to do to be qualified as "intelligent"? These questions relate to the next generation "intelligent" machines. Probably, these machines should be able to handle knowledge discovery, decision-making and concepts. In this paper, we will take into account some historical contributions and discuss these different questions through an analogy to human intelligence. Also, a general framework for a concept oriented language will be proposed.

## 1. Introduction

In 1961, Minsky[1] acknowledged that "only a few machines doing things might claim any real intellectual status". The question of *intelligent machines* was also posed a long time ago by Turing[2]. This latter also proposed what is now called the Turing test, in the same article. More recently, Chomsky[3], used the term "mental properties" as we could say "chemical" or "physical" properties. I support this idea and will use the term Artificial Intelligence (*AI*) in this way. Moreover, we don't know what "intelligence" means exactly. So, how could we define an "artificial" intelligence? In agreement with Chomsky, I will write about "learning machines" rather than "machine learning", and "decision machines" rather than "machine decision" (by the way, we could argue that this property, the ability to decide, could be a proof of intelligence, at least this is the point of view of some religions).

Through human history, mankind has been involved in nature in different ways: in the Neolithic period, people started to use tools to act with Nature, showing an intelligent property if we compare to other beings (see Figure 1). Then, from the industrial revolution, mankind began to use moving machines (motors, turbines, cylinders, etc.) and then memorizing machines (cams, gears, punch cards, etc.) with mechanical, (analog) electronic and (digital) electronic devices. We are, nowadays, at the age of data revolution and nearly all professions use or will use memorizing and learning machines (first based on coils and lamps and then, mainly on transistors). The next step is up to everyone to imagine. For me, the next step is a conceptual revolution in which discovering and decision machines would spread in all human activities. For this reason, I try to lay down the foundations of a concept oriented language able to handle such activities. By lack of time for coding, my proposal would be more philosophical than practical.

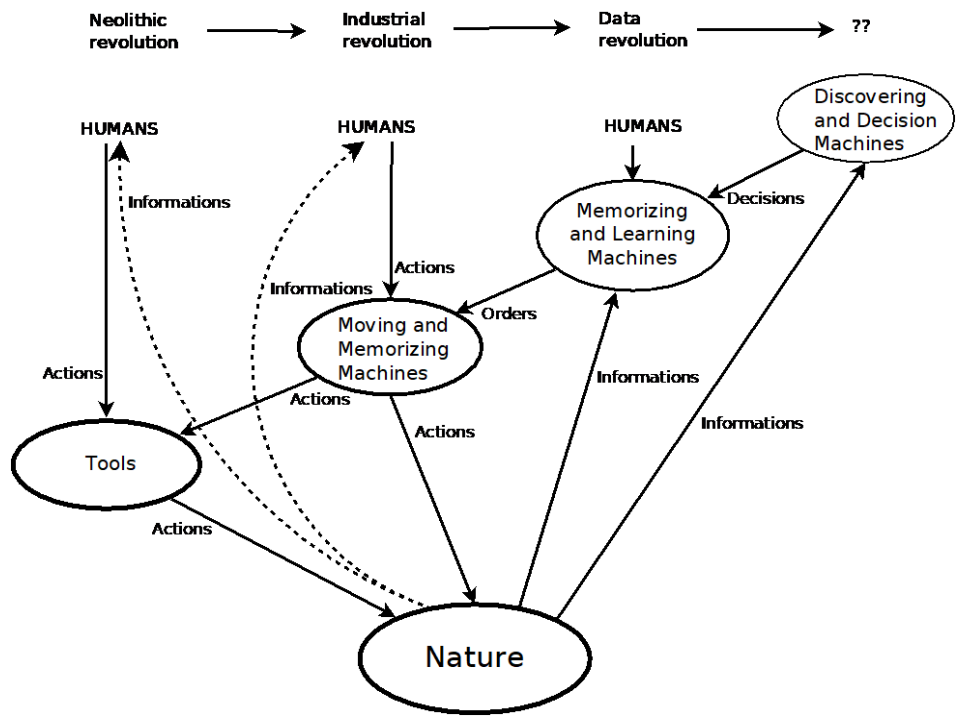

**Figure 1: evolution of the relationships between humans and Nature: first through tools, then through moving machines, after that through learning machines (data era) and, finally, we could expect through discovering and decision machines.**

The definition of mental properties could nowadays be updated considering all the evolutions since Turing's article. We should always keep in mind that planes have been invented by studying nature and specifically, beings that are able to fly. In the same way, we can design real intelligent systems or imitate some of the mental properties only if we study and keep in mind the *modus operandi* of brains.

We can notice, among others, the following phenomena from children's behavior:
- A child can learn by three main operations modes: by heart (memorizing), by thinking and by discovering through practice. Each operation mode can have different sub-modes (equivalent to algorithms) and each sub-mode creates its modeling framework and its models. It does not seem that all children use the same sub-modes.
- If a new object (meaning an instance of a class or pattern, in this case), is shown to a child, typically, he would ask "what's this, what is it for?". He needs to know which criterion he has to use in order to not confuse it with things he knows. For example, if it's a new variety of apple, he will need sometimes new criterions to distinguish it from other varieties (e.g. he could need the usual color of this variety).
- In our complex world, the amount of available data does not allow us to memorize every measure of everything. So, children need to use stereotypes or generalizations which are analogue to prototypes, i.e. models in learning machines.
- Also we can teach in different ways, give importance to one or the other at different times: technically, we can show all things (in a sense similar to instances of patterns) in each category (similar to patterns or models) and tell him to which category they all belong, or we can give different things from different categories, successively and repeat until all things are shown. These are the two main teaching strategies for a child.
- Finally, a child can also improve his knowledge by learning from others: communication, or information science (see, for example Shannon[4]), is inherent to learning.

In this paper, we will not treat all the involved learning paradigms, but propose a general framework and a concept oriented language as a next step in designing learning machines. Also, the related work is not focused on the more recent work in *AI*, since the problems are posed since at least 60 years. It is of great importance that machines have realistic models and classification paradigms (i.e.

techniques, algorithms) for the targeted area of research. We will use standards, prototypes and models as having the same meaning, and features, coordinates, parameters, variables, criterions and attributes as having the same meaning, for a unified approach to learning machines.

Intelligence is not only the ability to memorize a succession of procedures (with basic data or statistical data). For instance, PLCs (Programmable Logic Controllers) are not considered as *learning machines*: they behave according to what we have introduced in their memories (a set of logical rules, a set of sequences and a set of data), adaptively and dynamically. However, these are typical examples of machines doing only memorizations. Nowadays, we have intelligent machines which are able of different operations involving learning:

- Learning (dynamically).
- Discovering knowledge, for example, a clustering (see e.g. Dzeroski[5]) algorithm which tries to determine automatically the number of clusters or categories among a data set.
- Combining decisions from different machines, or classifiers.
- Making data sets redundant (Bagging and Bootstrap aggregating).
- Using incremental techniques to fit dynamically the model of knowledge. This is one of the operations modes useful for discovering machines.

Furthermore, since Turing, Human Science has evolved through modern mathematics which gave slowly birth of another form of computer science, which is now more and more practiced through object-oriented programming. More recently, some Frame Representation Languages[6] have been proposed and are used, for example in semantic web. We think that these are more recent attempts to define systems with associations of ideas, as in the early ages of *AI*. We propose to introduce a *concept oriented language* (COL) in representing knowledge, by formalizing a new general framework.

Such a language should use *Data Science* because we have much more data ("Big Data" is often hinted) available than before, and *Information Science*[5], because these data or induced data can and should propagate inside a machine and between different machines (this is analogue to humans neurons able to transmit information to other neurons and also able to be activated through communications with other brains). Data science can be described as a mixture of Mathematics (including statistics, probabilities and data analysis) and Computer Science. For a definition of Information Science, the reader can see, for example,

Shannon[4] and Zadeh[22]. Also, no measure of intelligence can be done without communication, i.e. information science, and this seems obvious in the Turing test. It is possible to introduce meta-vectors involving numerical and conceptual features (i.e. attributes or descriptors or criteria). This will involve statistical and structural methods.

In this paper, a brief taxonomy of existing intelligent machines and the involved research and application fields will be introduced. Then, a proposal for a concept oriented language (COL) will be discussed. After that, a definition of spaces, a framework for a COL and a case study will be introduced. Finally, the last section includes a conclusion and some prospects.

## 2. A taxonomy of existing intelligent machines

Many mental properties are implemented, through learning, clustering and updating machines. These machines use different algorithms or paradigms. These are able of:

- learning (many operation sub-modes of machine learning algorithms, as in Hastie et al.[7]):
    - Rule-based systems.
    - Knowledge-based systems
- clustering and categorization (see, e.g. Dzeroski[5]):
    - K-means, Fuzzy K-means examples (Chen and Ho[8]).
    - Other unsupervised (or semi-supervised) algorithms.
- mixing the two precedent categories (supervised and unsupervised learning), but without being able to change and add models:
    - Cooperation methods such as 3C algorithm (see, e.g. Gunes et al.[9]).
    - Dynamic ensemble selection (as in Ko et al.[10])
- evolving models but without changing or adding models :
    - Incremental methods (see, e.g. Neal and Hinton[11]).
    - Bagging, Boosting (see, e.g. Quinlan[12]).
- changing models but without evolving or adding models:
    - Selection based systems (see, e.g. Gunes et al.[9]).
    - Adaptive and/or dynamic selection of classifiers (see, e.g. Gunes et al.[13]).
    - Combining multiple classifiers (see, e.g. Gunes et al.[14], Bahler and Navarro[15], Lam[16]).

## 3. Involved research fields and technologies

Nowadays, learning machines and *DIS* systems are involved in many research fields and applied technologies. A list of them is shown in Figure 2, where the central circle represents the whole *learning machines* field. This field can be better described with *Data and Information Science*, which is the base for learning machines.

This list of fields (research and applied) is not exhaustive. Many algorithms and techniques can be applied in different if not all the applied fields. But in practice, due to habits and knowledge of the community working on learning machines, some applied fields are more associated with some of the research fields. These associations are shown in Figure 2.

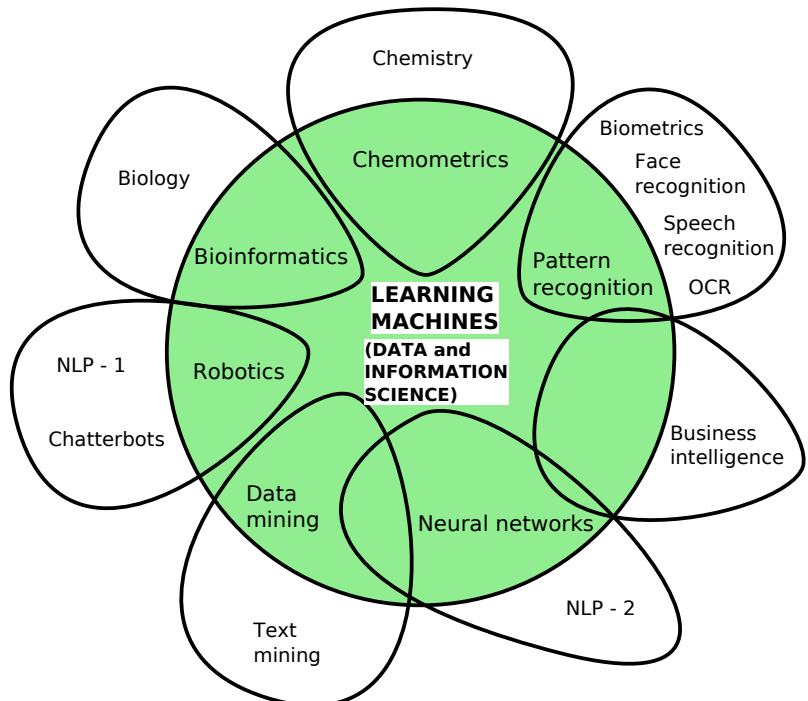

**Figure 2: non-exhaustive fields related to learning machines. The circle includes modern learning machines or *AI*. Some application fields can involve different areas of reseach fields and vice-versa. The overlappings are just shown as possible intersections between the fields. NLP is Natural Language Processing, OCR is Optical Character Recognition.**

## 4. A concept oriented language

We suggest to design a knowledge-based system (KBS) using a concept oriented language, such as the system can grow with the creation of new concepts and, of course, new instances (or classes) of these concepts. Minsky, who has been a student of Chomsky, has proposed a "Frame representation language" which is a knowledge-based system with reciprocal rules. Some (general) frame languages are KRL, OWL, LOOO. With these languages, the web can be organized (shown) by concepts in an ontology. We propose to define a KBS with a set of concepts and a set of frames (transformers). It should be able to take into account sequential orders of construction, and sequential learning. A correction of data, in an expert mode, could be useful.

### 4.1 An example of a KBS

In our sense, a concept can be represented (not exclusively) by a classifier and, if needed, can be completed by other classifiers as new features are created, in an incremental way.

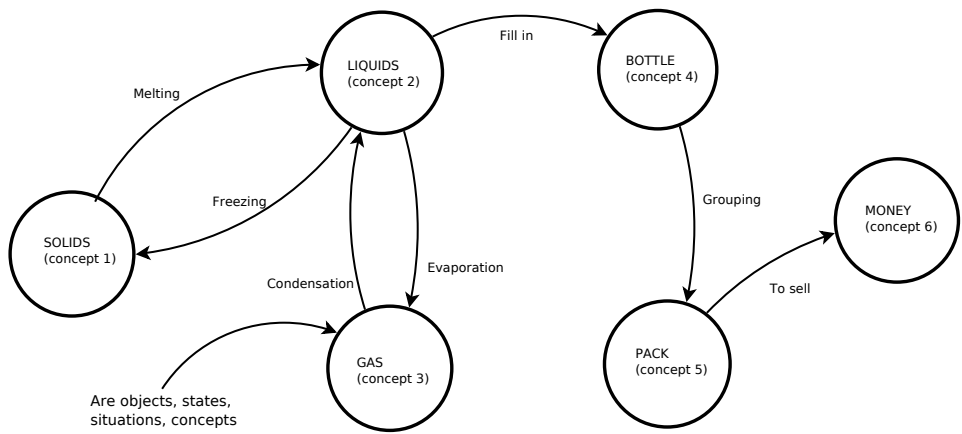

**Figure 3: example of concepts and frames allowing "transformations" from one concept to another.**

### 4.2 A proposal for a COL

The frames are inspired from psychology. They allow analogical reasoning, which is highly needed. This analogy can be allowed with features allowing a selection of a category of classes (in the Pattern Recognition or *PR* sense). They are similar to "relations" in the entity-relations model, which creates an ontology for a specific domain, but this model is rather adapted to Managing/marketing.

Some questions should be answered:
- how many "frames" (or « transformers »)  do we need ? As many as we need relationships between concepts. Maybe more, it's an open problem.
- which representation to choose ? We propose a representation which allows introducing concepts, frames and features, where adjectives would be represented by features such as follows:

|  | Entities-relations: | Our proposal: |
|---|---|---|
| Verbs | Relations | Frames |
| Subjects, objects | Entities | Concepts |
| Adjectives (quantitative, qualitative) |  | Features<br>One feature → one classifier |

In general, for a new object, the KBS should create a new concept, for a new verb, the system should create a new frame and with a new adjective we should create a new feature.

Such a Knowledge-based system should involve:
- at least, one classifier per concept ($n$ features need $n$ classifiers),
- a set of classes, in the pattern recognition (PR) sense, for each classifier,
- a set of frames (with oriented arcs which give causality, consistently with Pearl[23]), with input and output features among all incrementally created features and values for each of them (data associated with a concept).

## 5. An example of frame

An example of frame, the "evaporation" frame is shown below (see Figure 4). The set of rules linking one concept to another should be defined, as much as possible, as reciprocal rules. This will allow a reliable query of the KBS, when its time to exploit the KBS. But, non reciprocal rules can be allowed. In these rules, we could use qualitative features and quantitative features (in this case, we can use mathematical formulas). For this example we will need some one-sided rules such as (deduced from the *ideal gas law*: PV=n R T):

- If *n* and *V* are given and V≠0 then $P = \dfrac{n\ R\ T}{V}$

- If $n$ and $P$ are given and $P \neq 0$ then $P = \dfrac{n\ R\ T}{P}$
- If $P$ and $V$ are given and ($R \neq 0$, $T \neq 0$) then $n = \dfrac{P\ V}{R\ T}$

This can be improved, for example, by allowing these rules only for low pressures. Of course, these rules are activated only if the external temperature $T$ is given.

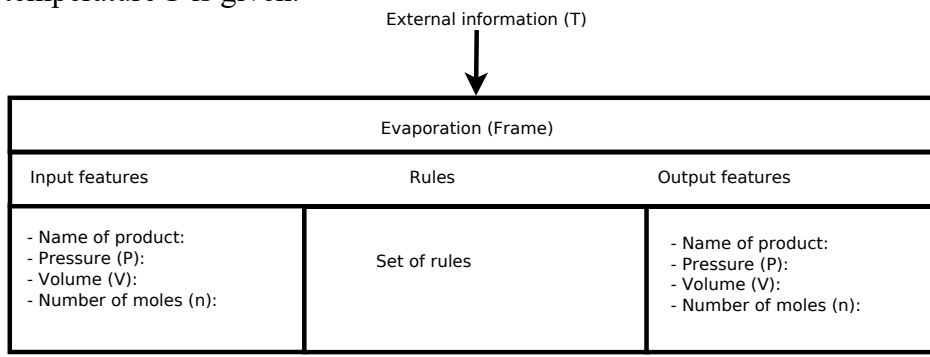

**Figure 4: a fundamental example of a frame (evaporation frame) which shows how we could define its input and output features, its set of rules and its optional external information (in this case, the temperature *T*).**

## 6. A definition of spaces

Let $F = \{f_1, f_2, \ldots, f_i, \ldots, f_j\}$ be the feature space in which the $f_i$ are different features and $j$ is the number of available features. We will need a set of classifiers $E = \{e_1, e_2, \ldots, e_k, \ldots, e_l\}$ to handle all the features. Let $\Omega = \{C_1, C_2, \ldots, C_m, \ldots, C_n\}$ be the decision space (also called space of discernment), in which the $C_m$ are the different classes or assumptions and $n$ is the number of classes at a moment. Thus, $n$ could increase or decrease incrementally.

According to the situation, a learning machine should be able to undertake:
- Learning, this should be a ubiquitous action.
- Categorization or Clustering (possibly creating or suppressing categories).
- Updating itself (discovering): by using incremental methods, discovering/suppressing classes, discovering the need for new features (incrementally), suppressing features. We could use the informal term of "evolved" models or classes, to detect if the models have changed enough to update the machine. Each system should have its own way of evolution detection. We can cite, however, the example of the information entropy (see Shannon[4]).

- Combining its decisions and/or decisions from outside of the machine.
- Giving an approximate answer when data is incomplete.

Let $\Delta = \{K_1, K_2, \ldots, K_o, \ldots, K_p\}$ be the concept space, in which the $K_o$ are the different concepts and $p$ is the number of concepts at a moment. Thus, $p$ can increase (or even decrease ?) incrementally, during learning.

## 7. A framework for a COL

Minsky [1] wrote "A computer can do, in a sense, only what it is told to do. But even when we do not know exactly how to solve a certain problem, we may program a machine to search through some large space of solution attempts". Also, he wrote "Certainly we must use whatever we know in advance to guide the trial generator. And we must also be able to make use of results obtained along the way."

The KBS should be able to add a new couple classifier/feature, each classifier being specialized with only one feature. This would help in simplification and generalization of the tasks (decisions, combination of results, etc.). Such a classifier could be based, for example, on one histogram for a unsupervised learning and on n histograms for a supervised learning approach, n being, in this case, the number of classes involved.

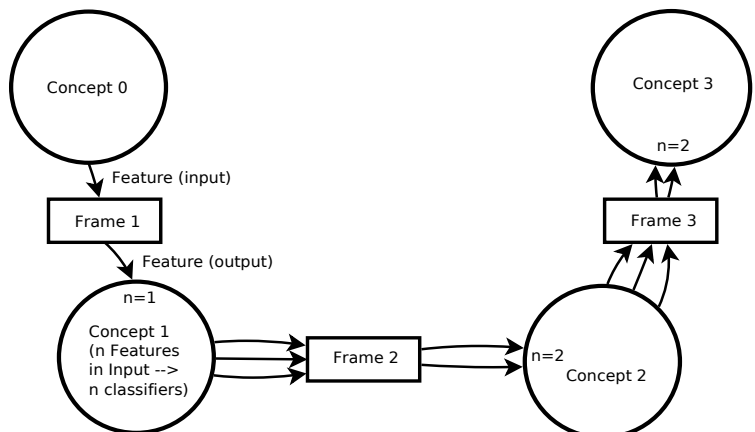

**Figure 5: a framework for a COL where arcs represents input features involved in respective frames. *n* is the number of features in input . For each concept, *n* is also the number of classifiers involved.**

## 8. A case study

As I emphasized on the shortcomings of the Turing test in a paper (Gunes et al.[24]), the following example is for natural language processing, but we could imagine different KBS in different fields using this approach.

Usually, objects can be separated ultimately by one and only one descriptive feature (which is also a *unique-concept,* denoted below as *u-concept*). We should, first, create three dictionaries:

- The fist for unique-concepts, such as: Breakable, Reliable, Hot, Cold, etc.,
- the second for concepts, such as: Possessions, Pains, Illness, Human, etc.,
- and the third for features' definitions, such as: Breakable: No, Yes, etc. ; Comfortable: No, Yes, etc. ; Expensive: No, Yes, etc.

If the trainer client uses/introduces a unique-concept then the KBS creates an instance of a feature of the corresponding unique-concept. An example of this kind of decomposition, for "Possessions", is shown below:

```
Possessions→  Glasses →Material     →Breakable (u-concept→feature):
(concept)     (concept)(concept)        No, Yes, etc.
                       →Shape (u-concept→feature):
                          Oval, Rectangular, Round, etc.

              →Shoes    →Material    →Comfortable (u-concept→feature):
                                        Yes, No, etc.
                                     →Expensive (u-concept→feature):
                                        No, Yes, etc.
```

**Hypothesis:** As a baby first knows "Dad", "Mum", he knows the class "Humans". We suppose that our machine knows at least the "Human" concept and some basic other concepts and frames ("Breakable" included).

**Beginning:** "Humans" and "Breakable" are concepts created in this KBS. Alan is a person training the system. Operations which are done by the system of KBS are shown in brackets. Our comments are written in italics. I means Inputs whereas O means Outputs.

1- Alan: **My glasses are broken !**
2- Machine (M**):** [Glasses: new class ? first data of a class ?
   "Glasses" do not exist: new class of the current concept (Humans)
   "Glasses" exists: no operation]

    **Your glasses ?**
3- Alan: **Yes, they felt down on stones. I can't <u>see</u> very <u>well</u> without them.**
4- M: [New concept: "See well"
New frame: "TO SEE"→ frame from "Humans" to "See well", see Figure 5.
I: Owns glasses: Yes, No
O: Quality: Good, Bad
Rule (example): If Humans own glasses, the quality of vision= good, else bad].
*For this level of learning, we would consider these rules as reciprocal, even if, obviously, they aren't.*
    **They can break easily ?**
5- Alan: **Well, I had mineral glasses, not synthetic ones** [2 qualitative adjectives→ values of different types of glasses]
*With quantifying adjectives we could create a Feature for each adjective.*
6- M: [I: Type of material: mineral, synthetic. Creates a feature of the class "Glasses" and a frame from "Humans" to "Breakable".
New frame: "To let fall" from "Humans" to "Breakable"
I: Type of material: Mineral, Synthetic
O: Breakable: No, Yes
Rule (example): If Type of material=Mineral then Breakable=Yes]
    **Why do you need glasses ?**
7- Alan: **If I don't use glasses, I have pain at my eyes.**
8- M: [New concept: "Pain at eyes"
New frame: "TO USE" from "See well" to "Pain at eyes"
I: Quality: Bad, Good
O: Pain at eyes: No, Yes
Rule (example): If Quality vision=Good then Pain at eyes= No, else Yes]

So after stage 2, we will have the following graph:

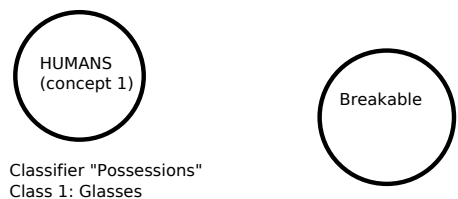

**Figure 6: graph obtained at stage 2.**

After stage 4, we will have:

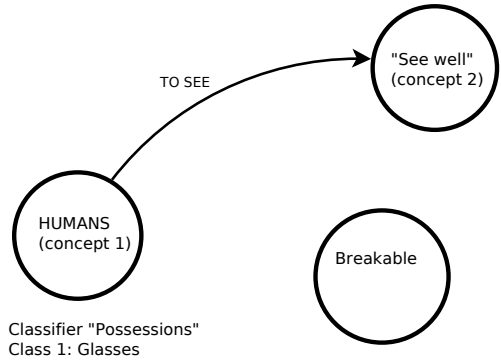

**Figure 7: graph obtained at stage 4.**

and a frame "TO SEE", such as:

| TO SEE: Humans to "See well" (frame) | | |
|---|---|---|
| Input | Rules | Output |
| - Owns Glasses: Yes, No | If Owns glasses=Yes ⇔ Quality vision=Good<br><br>If Owns glasses=No ⇔ Quality vision=Bad | - Quality vision: Good, Bad, … |

**Figure 8: frame created at stage 4.**

Then, after stage 6, we will have:

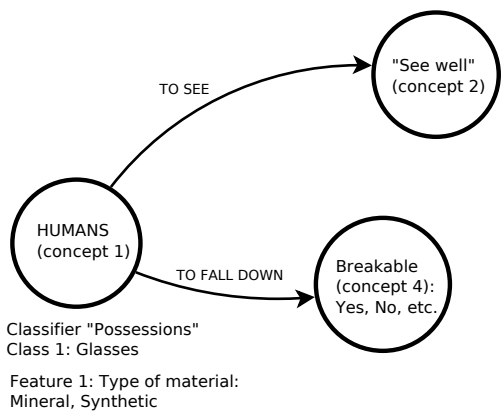

**Figure 9: graph obtained at stage 6.**

and a frame, "To let fall", such as:

| TO LET FALL: Humans to "Breakable" (frame) |||
|---|---|---|
| Input | Rules | Output |
| - Type of material: Mineral, Synthetic | If Type of material=Mineral ⟺ Breakable=Yes <br><br> If Type of material=Synthetic ⟺ Breakable=No | - Breakable: Yes, No, ... |

**Figure 10: frame created at stage 6.**

Finally, after discussions (after 8), we would have another frame: to use, such as:

| TO USE (Frame) |||
|---|---|---|
| Input | Rules | Output |
| - Quality vision: Bad, Good, ... | Quality vision=Good ⟺ Pain at eyes=No <br><br> Quality vision=Bad ⟺ Pain at eyes=Yes | - Pain at eyes: No, Yes, ... |

**Figure 11: frame created at stage 8.**

All rules will be in this form. Finally, we would obtain the following concept oriented graph:

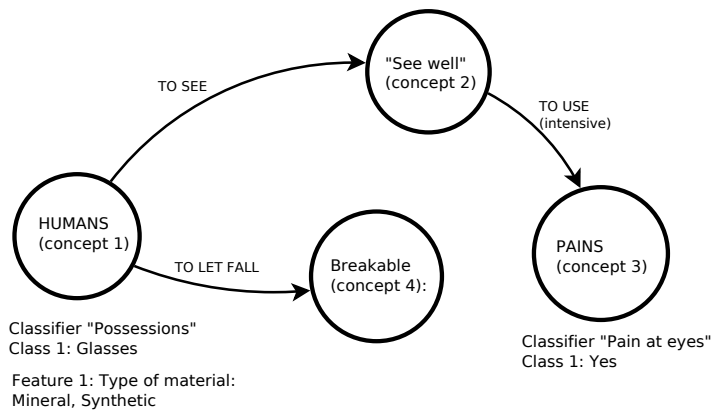

**Figure 12: final obtained graph.**

## 9. Test of the KBS

In order to check a correct learning, we can undertake a test such as:

Jack : Oh, my eyes are tired !
M: May be you have a bad vision ? [deduction from "to use" frame]
Jack: What should I do then ? (for a good vision)
M: You should wear glasses (deduction from "to see" frame)

Today, most of the learning machines are "tools" allowing to help us to categorize, classify, organize, study similarities, find evolution laws, collect information about a subject or a market, etc. To go further and check an evolving of the learning, we can ask the same questions (after another interrogator) and then check if the answers have evolved. This, added to the Turing[2] test, would significantly improve it, because if all the answers are the same, we can easily guess this as a machine (with limited learning).

## 10. Conclusion

An evolution of a knowledge base (real-time acquisition and introduction of new patterns into the base) is not enough to qualify a system as intelligent or a learning machine. Nowadays, machine learning should be able to do conceptual comparisons (as opposed to only numerical comparisons), for example: detect a fruit which is sweeter than others. It should be able of revolutions, such as to decide "this is a new concept and inside a concept, this is a new class". For example, it should decide: this is a new type of illness (unseen or unknown before).

In other words, these machines should be able to modify their knowledge base, such as to be able to modify their classes or concepts or add new classes or concepts. Causality should be explicit (in our example, with directed arcs) in the construction of a KBS, such a way that a class in a concept could allow deducing its cause from another concept.

We plan to use the proposed framework in the field of pattern recognition for learning problems where some features are numerical and others are symbolic and/or conceptual by nature. Proposals for collaborations, to show that the developed approach is possible, are welcome.